\pdfoutput=1

\documentclass[11pt]{article}

\usepackage[]{EMNLP2022}

\usepackage{times}
\usepackage{latexsym}
\usepackage{graphicx}
\usepackage{amsmath}
\usepackage{amssymb}
\usepackage{textcomp}
\usepackage{multirow}
\usepackage{array}
\usepackage{inconsolata}

\usepackage{tabularx, booktabs}
\newcolumntype{Y}{>{\centering\arraybackslash}X}
\newcolumntype{L}{>{\arraybackslash}X}

\usepackage[T1]{fontenc}

\usepackage[utf8]{inputenc}

\usepackage{microtype}

\usepackage{xparse}

\definecolor{bluepigment}{rgb}{0.2, 0.2, 0.6}

\newcommand{\ours}{P4E}

\title{P4E: Few-Shot Event Detection as \\ Prompt-Guided Identification and Localization}

\author{Sha Li$^1$, Liyuan Liu$^2$, Yiqing Xie$^3$, Heng Ji$^1$, Jiawei Han$^1$ \\
$^1$University of Illinois at Urbana-Champaign \\
$^2$Microsoft Research \\
$^3$Carnegie Mellon University \\
\texttt{\{shal2, hengji, hanj\}@illinois.edu} \\
\texttt{ lucliu@microsoft.com} \\
\texttt{yiqingxi@andrew.cmu.edu} 
}

\begin{document}
\maketitle
\begin{abstract}

We propose \ours, an identify-and-localize event detection framework that integrates the best of few-shot prompting and structured prediction.
Our framework decomposes event detection into an\textit{identification} task and a \textit{localization} task. 
For the identification task, which we formulate as multi-label classification, we leverage cloze-based prompting to align our objective with the pre-training task of language models, allowing our model to quickly adapt to new event types. 
We then employ an event type-agnostic sequence labeling model to localize the event trigger conditioned on the identification output.
This heterogeneous model design allows \ours~to quickly learn new event types without sacrificing the ability to make structured predictions.
Our experiments demonstrate the effectiveness of our proposed design, and \ours~shows superior performance for few-shot event detection on benchmark datasets FewEvent and MAVEN and comparable performance to SOTA for fully-supervised event detection on ACE. 

\end{abstract}

\section{Introduction}
Event detection is an essential step in information extraction, 
which aims to locate the event trigger (i.e., the minimal lexical unit that indicates an event) and classify the event trigger into one of the given event types.
While steady progress has been made
for event detection given ample supervision~\cite{wadden-etal-2019-entity,lin-etal-2020-joint,lu-etal-2021-text2event}, it is hard to replicate these success stories in new domains and on new event types, due to their reliance on large-scale annotation.
\begin{figure}
    \centering
   \includegraphics[width=\linewidth]{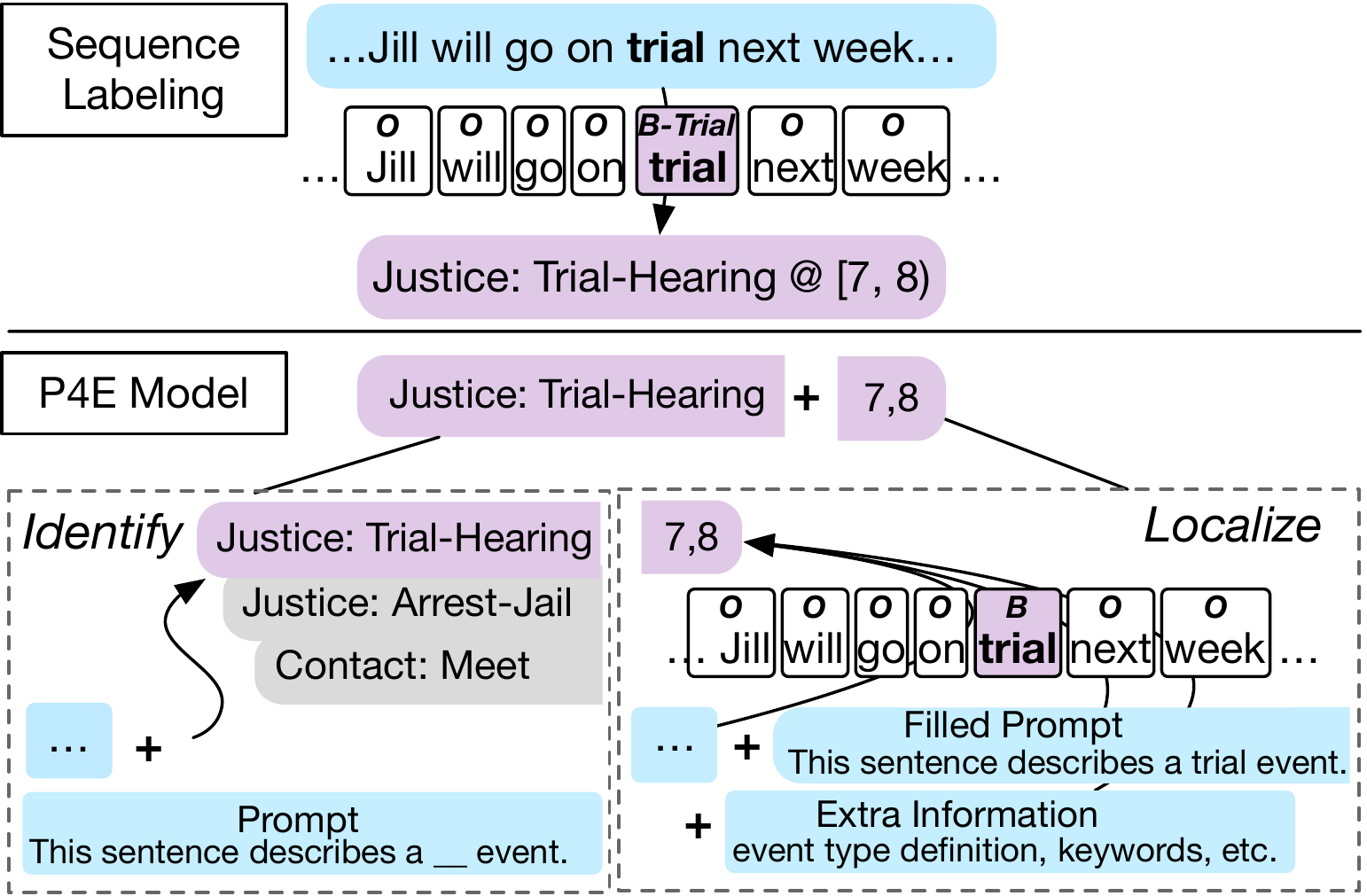}
    \caption{
    Event detection requires the model to produce both event types and trigger locations. In our  \ours~model, we decompose the task into two stages of identification and localization to take advantage of few-shot prompting and reuse the structured prediction model.
    }
    \label{fig:intro}
\end{figure}

Instead of asking humans to exhaustively annotate hundreds, or even thousands, of documents, a more economic and realistic setting is to ask for a handful of labeled examples per type, namely the few-shot learning setting.
Recently, prompt-tuning has shown great success in few-shot learning for a range of classification and generation tasks, thanks to better alignment between the target task and the language model's pretraining objective.
In the context of event detection, we could use a prompt such as ``The sentence describes a \texttt{[MASK]} event'' to guide the language model towards identification of the event type mentioned in the context.

However, since event detection requires recognizing both the event type and the trigger location,
the aforementioned cloze-based prompt learning paradigm~\cite{schick-schutze-2021-exploiting} does not give us the full solution.
Hence, we propose to decompose the event detection task into the identification task and the localization task.
Specifically, we first recognize the event types in the given context (the \textit{identification} stage) using  multi-label classification 
and then find the trigger location (the \textit{localization} stage) through structured prediction.

\begin{figure*}[t]
    \centering
    \includegraphics[width=.8\linewidth]{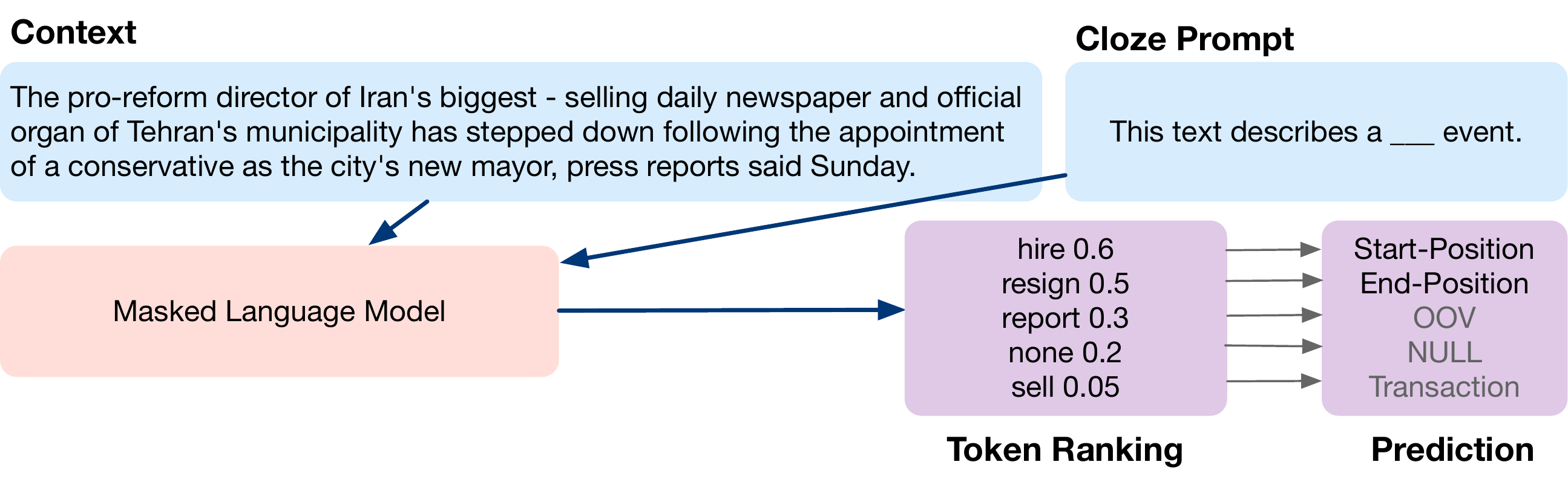}
    \caption{The identification model. The context and cloze prompt are concatenated and provided as input to the masked language model (MLM). The MLM produces scores for every token in the vocabulary as a measure of how well the token fits into the ``blank".
    Some tokens in the vocabulary can be mapped back to event types, such as \texttt{hire} $\rightarrow$ \texttt{Start-Position}. If a token does not map to any event type in the ontology (e.g., \texttt{report}), it will be ignored. 
    We predict all event types that have a higher score than the NULL label (which maps to the token \texttt{none}).
    }
    \label{fig:id_model}
\end{figure*}

Our identification model extends cloze-based prompt learning~\cite{schick-schutze-2021-exploiting} to multi-label classification.
Since a sentence can contain multiple events or no events at all, we extend the model to a multi-label classification setting by adding a NULL class which stands for ``no event identified''. 
We designate a special token \texttt{none} as the verbalizer\footnote{The verbalizer is a mapping from the class label to a single token in the language model's vocabulary.} for the NULL class and compare it against the predictions for all of the valid event types (as in Figure~\ref{fig:id_model}). 
In this design, the NULL verbalizer effectively serves as the dynamic threshold 
for multi-class classification~\cite{Zhou2021AdaptiveThreshold}. 

As to the localization model, 
we integrate the filled prompt along with optional event type descriptions and keywords\footnote{For the ACE dataset, we referred to the annotation guidelines, for MAVEN we used FrameNet. } to augment the input, and employ a sequence tagger to recognize the event trigger (e.g., ``appointment'' in Figure~\ref{fig:loc_model}).
In this way, we decouple the parameters of the localization model from the event label by including the event type information on the input side instead. %

Our experiments show that \ours~achieves the new state of the art for few-shot event detection across two benchmarks, i.e., FewEvent~\cite{Deng2020MetaLearningWD} and  MAVEN~\cite{wang-etal-2020-maven}. 
To further verify the effectiveness of our integration, we also test our model on the fully-supervised event detection benchmark  (ACE2005), and observe that \ours~ is comparable to the state-of-the-art performance. 

\vspace{.2cm}
Our study has three major contributions.
\vspace{-.2cm}
\begin{itemize}
\setlength\itemsep{-0.15em}
    \item We propose to decompose the event detection task into stages of identification and localization. By decoupling the type semantics from the sequence labeling task, we bring the benefits of cloze-based prompt learning to event detection and enable maximal parameter-sharing across event types, greatly improving the data-efficiency of our model. 
    
    \item  %
    We extend the cloze-based prompt learning paradigm to multi-label event type classification. 
    This enables us to leverage the language modeling ability of pretrained LMs for the event identification task and adapt quickly to new event types. 
    This method can be applied to other multi-label classification tasks.
    \item Our model achieves excellent performance on the event detection task under both few-shot and fully-supervised settings. In particular, for few-shot event detection on FewEvent~\cite{Deng2020MetaLearningWD}, we outperform the next best baseline by 15\% F1. On MAVEN~\cite{wang-etal-2020-maven}, we achieve 3\% gains in the overall event detection task.
\end{itemize}

\section{Methodology}

Given a collection of contexts $\mathcal{C}$ and a pre-defined event ontology $\mathcal{T}$ (a set of target event types), event detection aims to find all event mentions in the collection that fall into the given ontology.
An event mention is characterized by a trigger span $s$ (start index, end index)  and an event type $t \in \mathcal{T}$. Here we follow previous work and consider each sentence as the context of the event mention.

We divide the event detection task into two stages: identification and localization. 
In the identification stage, for each context $c$, we find a set of event types $T$ that have been mentioned. In the localization stage, we take a pair of context and event type $(c, t)$ as input and find a set of spans $S$ that correspond to the triggers for that event type.
Note that both stages can produce a variable number of outputs for each input: each sentence can contain multiple event types and each event type can have multiple triggers.

\subsection{Event Type Identification}
The event type identification model follows the idea of using a cloze-style prompt for few-shot learning with masked language models~\cite{schick-schutze-2021-exploiting}. 
Cloze-style prompt learning transforms a classification problem into a masked language modeling problem using a \textit{prompt} and a \textit{verbalizer} function.
The \textit{prompt} $P$ is a natural language sentence with a \texttt{[MASK]} token. 
Figure \ref{fig:id_model} shows a cloze prompt that can be used for event detection: ``This text describes a \underline{\texttt{[MASK]}} event". 
The relationship between the class labels $\mathcal{L}$ and the predicted tokens $V$ for the \texttt{[MASK]} is defined by the \textit{verbalizer} function $f_v$: $\mathcal{L} \rightarrow V$. For example, we choose the verbalizer function to map the event type \texttt{Start-Position} to the token \texttt{hire}. %

During prediction, we use the logit that the masked language model $M$ assigns to the verbalizer $f_v(l)$ for label $l$ as the proxy for predicting $l$. In the classification task, the probability for label $l$ can then be computed as shown in Equation \ref{eq:classification-prob}. 

\begin{equation}
\small
    p(t=l) = \frac{\exp \left( M(f_v(l)| x, P) \right) }{\sum_{l' \in \mathcal{L}}  \exp \left( M(f_v(l')| x, P) \right)}
    \label{eq:classification-prob}
\end{equation}

For event detection, since each sentence can potentially mention multiple event types, we extend this approach to handle multi-label classification.
Through the masked language model, we score all tokens in the vocabulary on their likelihood to fill in the blank. After excluding tokens that do not map back to any event type of interest (such as the token \texttt{report} in the example), we obtain a ranking among all event types. The key becomes finding the cutoff threshold for translating these scores into outputs. We assign a token $v_{\text{NULL}}$ to the NULL type\footnote{In our experiments, we use the token ``none" as the NULL type's verbalizer.} and use it as an adaptive threshold. In the inference stage, we predict all event types that score higher than the NULL type to be positive. In our example, since \texttt{hire} and \texttt{resign} both have higher scores than the NULL verbalizer \texttt{none}, we predict \texttt{Start-Position} and \texttt{End-Position} as the event types in the context.

During training, for each sentence, we compute the loss for the positive event types and the negative event types separately with respect to NULL:
\begin{equation}
\small
    \mathcal{L}_\text{pos} = \frac{1}{|T|} \sum_{t \in T} \log \frac{\exp (M(f_v(t) |x, P))}{
    \sum\limits_{t' \in \{ \text{\tiny NULL}, t\} } \exp (M(f_v(t') |x, P)) }
\label{eq:id-pos-loss}
\end{equation}
where $T$ is the set of positive event types for the sentence and $\bar{T}$ is the set of the remaining (negative) event types. Note that in computing the positive loss term, we only compare the probability of the positive labels to the NULL label. 

\begin{equation}
\small
\mathcal{L}_\text{neg} = \log \frac{\exp (M(v_{\text{\tiny NULL}}|x, P) }{ \sum\limits_{t' \in \{\text{\tiny NULL}\} \cup \bar{T}} \exp(M(f_v(t') |x, P)) }
\label{eq:id-neg-loss}
\end{equation}

\begin{equation}
    \small 
    \mathcal{L}_{id}  = \frac{1}{|C|} \sum_{c \in \mathcal{C}} \left( \mathcal{L}_\text{pos} +  \mathcal{L}_\text{neg} \right) 
\label{eq:id-loss}
\end{equation}

Equation \ref{eq:id-pos-loss} effectively pushes the score of each positive event type above NULL and Equation \ref{eq:id-neg-loss} lowers the scores for all negative event types.

For some event types such as "Business:Lay off", the natural language label ``lay off" cannot be mapped to a single token. In this case, we add a new token $\langle$\texttt{lay\_off}$\rangle$ 
and initialize its embeddings as the average of the tokens that compose the original event type name.

\begin{figure*}[t]
    \centering
    \includegraphics[width=.9\linewidth]{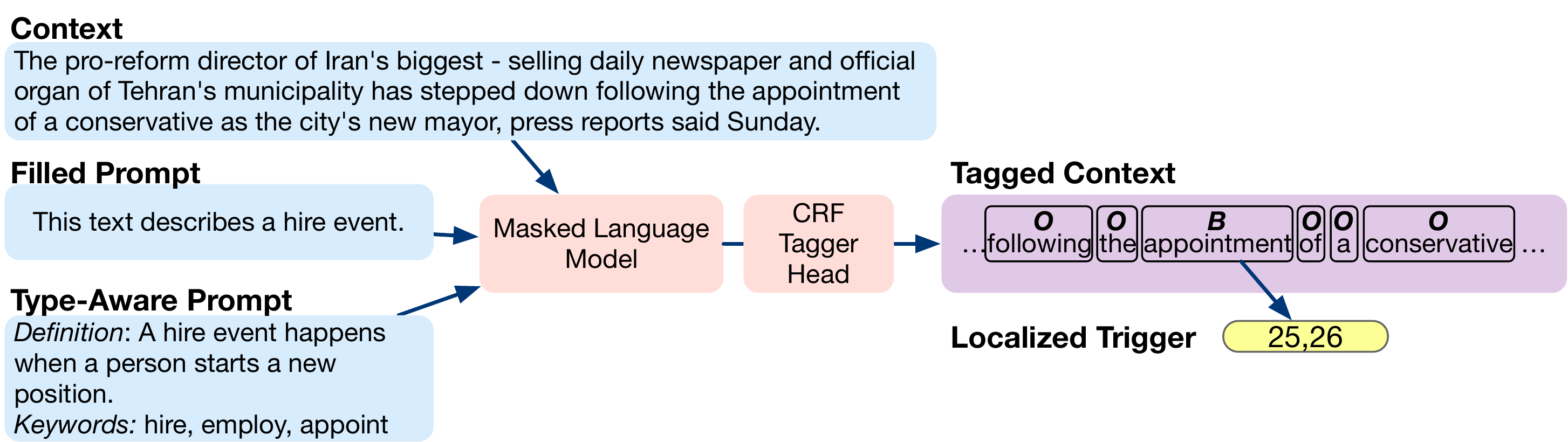}
    \caption{The localization model. The context, filled prompt (from the identification stage), and a \textit{type-aware prompt} are provided as input. The \textit{type-aware prompt} can be the event definition or event keywords. Our model outputs type-free BIO tags for the context which can then be converted into trigger locations.}
    \label{fig:loc_model}
\end{figure*}
\vspace{-5pt}
\subsection{Verbalizer Selection}
In the previous section we've assumed that either the event type label can directly serve as the verbalizer (\texttt{Attack} $\rightarrow$ attack)  or a manually picked verbalizer (\texttt{Start-Position} $\rightarrow$ hire) is provided. We will now describe a simple method for automatic selection of verbalizers in case neither of the two options is available. 

We first collect a candidate verbalizer set $\mathcal{V}$ from the training set examples' trigger words. Then we use a \textit{frozen language model} to score each candidate $v$ by plugging it in the cloze prompt of our identification model.
The selection for each event type $t$ is done separately. 
We compute the score of a candidate verbalizer $v$ for event type $t$ with class label $l_t$ using 
a reciprocal rank scoring function: 
\begin{equation}
\small 
    \text{score}(v, l_t) = \sum_i \frac{1}{r_i(v)}\mathbb{I}(y_i=t)
\end{equation} 
$r_i$ the predicted ranking of the candidate verbalizer from the frozen language model. We also experimented with the cross entropy scoring function but discovered that it favored more frequent words.
The verbalizer for each class is then selected to maximize the scores: $f_v(l) = \arg \max_{\mathcal{V}} \text{score}(v, l_t)$. 
Note that the surface form of the verbalizer is only an initialization and the word embedding of the verbalizer token will be updated during training. 

\subsection{Trigger Localization}
Trigger localization is the task of finding the trigger offset given a context $c$ and an event type $t$. Since we already know the event type, we can construct a more informative input by leveraging external knowledge (for instance, from FrameNet) about the event type.
For example, in Figure \ref{fig:loc_model}, we use the event description from the annotation guidelines to help define the ``Start-Position" event type. We can also use a few keywords (example triggers) to enrich the event representation. In our experiments, we compare these two forms of event knowledge.

Our localization model is a linear chain CRF tagger with only three tags: BIO\footnote{B stands for the beginning of a span, I stands for the inside of a span, and O for outside of any span. }.
In this way, the model parameters are not tied to any event type and can be easily used for transfer.

The probability of a tagged sequence is: 
\begin{equation}
\small 
    p(y | \vec{h} ; \theta) = \frac{\exp \left( \sum_i  \varphi(y_i|h_i)  + \sum_i \psi (y_i | y_{i-1}) \right)}{Z} 
\end{equation}
where $\vec{h}$ is the contextualized embedding vector of the tokens from the masked language model and $Z$ is a normalization factor.

We parameterize the emission scorer $\varphi(y_i|h_i)$ as:
\begin{equation}
    \varphi(y_i|h_i) = W_l h_i + \sum_j \alpha_{ij} W_v h_j 
\end{equation}
Both $W_l \in \mathbb{R}^{3 \times m}$ and $W_v \in \mathbb{R}^{3 \times m}$ map the embeddings to the tag space, serving as an early prediction.
Then we fuse the predictions for the token and the other tokens through an attention mechanism with the weight $\alpha_{ij}$ defined as:
\begin{equation}
\small 
    \alpha_{ij} = \text{Softmax}_j \left(\frac{(W_q h_i)^T W_k h_j }{\sqrt{m}}\right)
\end{equation}
$m$ is the dimension of the embeddings $h$ and $W_q \in \mathbb{R}^{m \times m}$, $W_k \in \mathbb{R}^{m \times m}$ are learnable parameters.

\subsection{Joint Training}
In a sense, our identification model captures the probability of the event type given the context $p(t | x)$ and our localization model captures the probability of the token tags given the context and event type $p(y | t, x)$.

The identification model and the localization model share the same masked language model backbone.
Since these two tasks have slightly different inputs, we alternate between sampling batches for each task.

\section{Experiments}

\paragraph{Datasets}
We evaluate our model on three datasets, FewEvent~\cite{Deng2020MetaLearningWD}, MAVEN~\cite{wang-etal-2020-maven} and ACE2005\footnote{\url{https://www.ldc.upenn.edu/collaborations/past-projects/ace}}.

We present the overall dataset statistics in Table \ref{tab:datasets}. Details of the datasets and data splits are available in the Appendix \ref{sec:dataset}.

\begin{table}[t]
    \centering
    \scalebox{0.9}{
    \small 
    \begin{tabular}{c|c c c c}
    \toprule 
    Dataset     & \# Docs & \# Sents & \# Event types & \# Instances \\
    \midrule 
    ACE+     & 599  & 20,818 & 33 & 5,311 \\
    FewEvent & - & 70,852 & 100 & 70,852 \\
    MAVEN & 4,480 & 49,873 & 168 &  118, 732 \\
    \bottomrule 
    \end{tabular}
    }
    \caption{Dataset statistics.}
    \label{tab:datasets}
\end{table}

\paragraph{Evaluation Metrics}
For all experiments, we use the event mention precision, recall and micro-F1 score as our major evaluation metrics~\cite{ji2008refining,lin-etal-2020-joint}. 
An event mention is considered correct if both its type and trigger span are correct.

\paragraph{Implementation Details}
We use BERT~\cite{devlin-etal-2019-bert}, specifically \texttt{bert-base-uncased}, for the experiments on FewEvent and MAVEN unless otherwise specified. For experiments on ACE, we used  Roberta~\cite{Liu2019RoBERTaAR}. 
For the base model, we use a batch size of 8 and a learning rate of $2e-5$. For the large model, we use a batch size of 16 and a learning rate of $1e-5$.
We set the maximum sequence length to 200 tokens since our predictions are on the sentence-level.
For more details, we refer the readers to the Appendix.

\begin{table*}[t]
    \centering
    \small 
    \begin{tabular}{c|l| c c }
    \toprule 
      Dataset   & Method & $K=5$ & $K=10$  \\
      \midrule 
     FewEvent (10 types)    &  BERT-CRF & 44.06 & 66.73 \\
     & PA-CRF\cite{cong-etal-2021-shot}*  & 58.48 & 61.64  \\
    & PA-CRF-Adapted & 63.64 & \underline{70.69} \\
    &      Prompt+QA & \underline{65.23} & 67.50\\
     &    \ours & 90.7 $\pm$ 2.2 /\textbf{81.98} $\pm$ 2.0 & 92.7 $\pm$ 1.1/\textbf{85.50} $\pm$ 1.3\\
     \midrule 
     MAVEN (45 types) & BERT-CRF & 48.14 & 64.68 \\ & StructShot~\cite{yang-katiyar-2020-simple} & 42.57 & 49.18\\
     & PA-CRF-Adapted & 53.16 &  \underline{65.62} \\
     & RelNet + Causal*~\cite{Chen2021HoneyOP}  & \underline{57.0} & - \\
     & Prompt + QA & 47.86 & 65.43 \\
      & \ours &  63.9 $\pm$ 0.9/\textbf{60.64} $\pm$ 1.0 & 72.6 $\pm$ 1.5/\textbf{69.51} $\pm$ 1.5 \\
     \bottomrule 
    \end{tabular}
    \caption{Few-shot event detection micro-F1 scores(\%) on FewEvent and MAVEN with different number $K$ of training examples per type. Models with * are taken from their original paper. For our model, the first number is the identification stage performance. We report the average performance and variance over 10 random seeds.  }
    \label{tab:fewshot}
\end{table*}

\begin{table*}[t]
    \centering
    \small 
    \scalebox{0.9}{
    \begin{tabularx}{1.1\linewidth}{m{24em}|YYYY}
    \toprule 
    \multirow{2}{*}{Context} &  \multicolumn{4}{c}{Model Predictions} \\
    \cmidrule{2-5} 
    & BERT-CRF & PA-CRF-Adapted & Prompt-QA & \ours  \\
    \midrule 
    Private donations covered one-third of the US
    \$20 million \underline{cost}[Cost] of the rescue, with the rest coming from the mine owners and the government. & Hiding\_objects: covered; 
    Cost: cost & Filling: covered; Cost: cost & Cost: cost & Cost:cost \\
    Indian Home Minister P. Chidambaram visited the state on Monday, 30 July to review the security situation and the relief and \underline{rehabilitation}[Cure] measures being taken. & - & - & Cure: rehabilitation & Cure: rehabilitation \\
    In December 1953, the British colonial government in Singapore passed the National Service Ordinance, \underline{requiring}[Obligation] all male British subjects ... to register for part-time National Service. & Rite: Service & Rite: Service; Submit\_Documents: register & Obligation: requiring & Obligation: requiring \\
    Its explosion created panic among local residents, and about 1,500 people were injured seriously enough to seek medical \underline{treatment}[Cure]. &  Cure: treatment & Institutionalize: medical; Cure: treatment &  Institutionalize: medical treatment; Cure: treatment & Cure: treatment \\
    \bottomrule 
    \end{tabularx}
    }
    \caption{Case studies on the few-shot event detection on MAVEN. The annotations are marked in the context: the trigger is underlined and the corresponding event name is provided in the bracket. 
    }
    \label{tab:fewshot_casestudy}
\end{table*}

\begin{figure}[t]
    \centering
    \includegraphics[width=\linewidth]{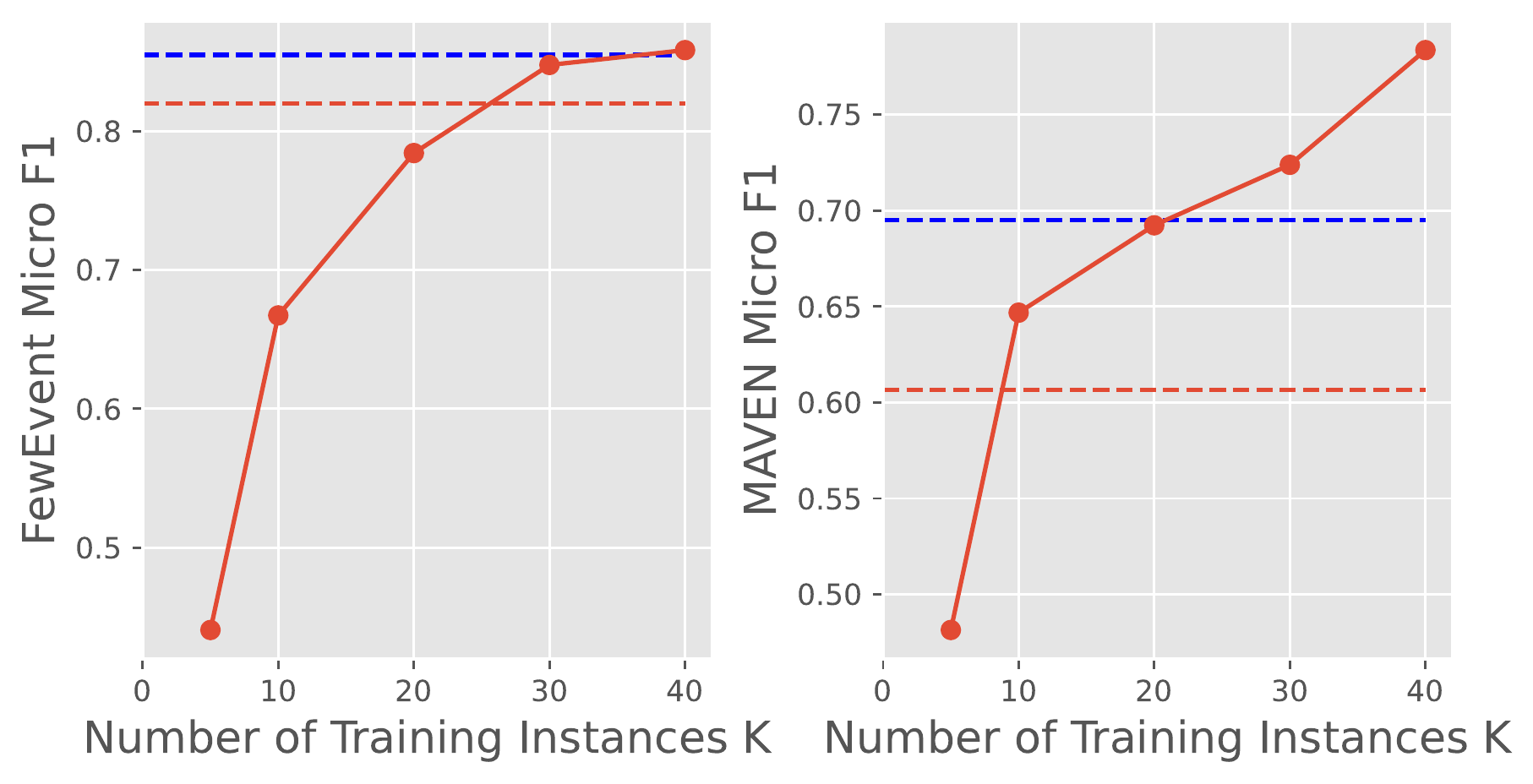}
    \caption{Comparison of our model with fine-tuning BERT-CRF on few-shot event detection. Red dashed line is \ours~ performance with $k=5$ and the blue dashed line is \ours~ with $K=10$.}
    \label{fig:compareK}
\end{figure}

\subsection{Few Shot Event Detection}
\paragraph{Experiment Setting}
For few-shot experiments, for \ours  ~and our implemented baselines, 
we follow the setting in ~\cite{yang-katiyar-2020-simple, Chen2021HoneyOP} which samples $K$ examples per type for training and uses the remaining samples for testing.  This is different from the episode-based setting in (1)  no access to extra event types for training; (2) the model has access to all of the $K$ examples for test types before inference. 
\paragraph{Baselines}
Our first baseline is the sequence classification model \textbf{BERT-CRF} which represents fine-tuning performance. Then we compare with three few-shot models \textbf{StructShot}~\cite{yang-katiyar-2020-simple}, \textbf{PA-CRF}~\cite{cong-etal-2021-shot} and \textbf{RelNet+Causal}~\cite{Chen2021HoneyOP}. 
In particular, PA-CRF was originally designed for the episode-based setting, so we implemented a variant \textbf{PA-CRF-Adapted} for our setting. Apart from the data difference, \textbf{PA-CRF-Adapted} stores the event type prototypes and recomputes them every epoch, making it more powerful than the original model, which computes prototypes from each batch and then discards them. \textbf{Prompt+QA} is a variant of our model by replacing the single-class CRF model with a QA scoring module.
\paragraph{Results}
We show our main results in Table \ref{tab:fewshot}, some sample predictions in Table \ref{tab:fewshot_casestudy} and list our findings as follows: 

\begin{itemize}
    \item In general, models perform better on FewEvent compared to MAVEN due to less number of event types to choose from. In fact, on FewEvent, our model's identification performance can reach 88.8\% for the 5 shot case, whereas on MAVEN the performance is only 63.9\%. The identification stage becomes the major bottleneck for our model.
    \item On the localization task, our model can jointly learn from annotation of all event types, giving us a significant advantage (over 15\% F1 on FewEvent) over models that rely on event type-specific parameters. 
    \item A common mistake of BERT-CRF and PA-CRF-Adapted is that the words are taken too literally and the model does not pay enough attention to context. As shown in the first example in Table \ref{tab:fewshot_casestudy}, \textit{covered} could be a trigger for \texttt{Hiding\_Objects}, but not in the context of \textit{covered costs}. They also do not work well on instances with low frequency triggers, such as \textit{rehabilitation}.
    \item Comparing the Prompt-QA variant and our model, we notice that the QA needs a larger total number of examples to be well-trained. %
    Since the head and tail are scored independently, when the number of examples is small, the QA model will often predict a long span instead of a trigger. With sufficient training examples, the QA model will perform relatively well (as in the K=10 case on MAVEN), but occasionally produce competing triggers such as \textit{medical treatment} and \textit{treatment} in the last example of Table \ref{tab:fewshot_casestudy}.
\end{itemize}

In Figure \ref{fig:compareK} we change the number of examples used for fine-tuning in comparison with our model's few-shot performance. On FewEvent, 
our 10-shot performance is nearly the same as fine-tuning with 40 examples, saving 75\% of the annotation. On MAVEN, due to the increased number of event types, the few-shot performance is not as compelling, but we can still save 50\% of the annotation cost when $K=10$.

\begin{table*}[ht]
\small 
    \centering
    \begin{tabular}{l | l|c c c }
    \toprule 
      Category &  Model  &  Prec & Recall & F1  \\
       \midrule 
       Sequence labeling & Token Classification & 67.1 & 72.3 & 69.6 \\
       Sequence labeling & Token Classification+CRF & 67.8 & 76.6 & 71.9 \\
       Multitask & OneIE*~\cite{lin-etal-2020-joint}  & - & - & 72.8 \\
       Multitask & FourIE*~\cite{nguyen-etal-2021-cross} & - & - & \underline{73.3} \\
       QA & EEQA*~\cite{du-cardie-2020-event} & 71.1 & 73.7 & 72.4  \\
       Generation & Text2Event*~\cite{lu-etal-2021-text2event} & 71.2 & 72.5 & 71.8 \\
       Generation & DEGREE*~\cite{hsu2021degree} & - & - & 72.7 \\
       Prompt-based & \ours & 70.9 & 76.1 & \textbf{73.4} \\
       \bottomrule 
    \end{tabular}
    \caption{Supervised event detection results (\%) on ACE+. The best results are in boldface and the next best results are underlined.* indicates results cited from the original paper.}
    \label{tab:ace-results}
\end{table*}

\subsection{Supervised Event Detection}

We report supervised event detection results on the ACE+ dataset in Table \ref{tab:ace-results}. 
We compare with a wide range of existing methods, covering the paradigms of single-task sequence labeling, multitask learning, question answering and generation. We see that our prompt-based task formulation performs equally or better than existing methods.
In particular, the multitask learning models such as OneIE~\cite{lin-etal-2020-joint, nguyen-etal-2021-cross} enjoy the benefits of joint training across related tasks such as entity extraction and relation extraction.
Notably, DEGREE~\cite{hsu2021degree} also uses event descriptions and keywords as a ``type-aware prompt" to guide the generation of the trigger word. However, generation using the entire vocabulary is more challenging than our localization task.

\section{Analysis and Discussion}
\subsection{Automatic VS Human Verbalizers}
\label{sec:verbalizer_choice}
We present some of the automatically selected verbalizers in Table \ref{tab:verbalizers}. In general, these verbalizers are not far off from the event type semantics, but may be ambiguous (such as the word ``house") or have more broad usage beyond the scope of the event type (such as ``design" may be used outside of art).

In Table \ref{tab:maven-autoverbalizer}, we show how this difference in verbalizers affects the event detection performance.
When the number of examples per event type increases, the verbalizer selection quality is improved and the model is also able to rely more on the training examples instead of the verbalizer initialization, leading to a smaller gap between the automatic selection and manual selection.

\begin{table}[th]
    \centering
    \small 
    \begin{tabular}{l|c c}
    \toprule 
       Event type   & Manual & Automatic  \\
       \midrule 
       Filling   & fill & cover \\
       Cure & treatment & relief \\
       Create\_artwork & draw & design \\
       Imposing\_obligation & require & charges \\
       Commerce\_buy & purchase & shopping \\
       Containing & contain & house \\
       \bottomrule 
    \end{tabular}
    \caption{Examples of automatically selected verbalizers when provided with 10 examples per class.}
    \label{tab:verbalizers}
\end{table}

\begin{table}[th]
    \centering
    \footnotesize
    \begin{tabular}{l|l c c }
    \toprule 
    Task     &  Model &  $K=5$ & $K=10$  \\
    \midrule 
      Id  
      & Automatic & 59.5 $\pm$ 1.5 & 70.4 $\pm$ 1.4 \\
      & Manual & 63.9 $\pm$ 0.9 & 72.6 $\pm$ 1.5 \\
      \midrule 
      Id + Loc
      & Automatic & 56.8 $\pm$ 1.2 & 67.5 $\pm$ 1.1 \\
      & Manual & 60.6 $\pm$ 1.0  & 69.5 $\pm$ 1.5 \\
      \bottomrule 
    \end{tabular}
    \caption{Few-shot event detection results (\%) on MAVEN with automatic selected verbalizers. }
    \label{tab:maven-autoverbalizer}
\end{table}

\subsection{Impact of NULL Instances}
Prior datasets on few-shot event detection regularly ignore the presence of NULL instances since they are composed of i.i.d. sampling over the candidate event type. We discuss this problem in detail in Appendix \ref{sec:appendix-dataset}. In real-life applications, the model will inevitably encounter sentences that do not contain any events of interest. To test our model's capability of handling such cases, we inject NULL instances into the MAVEN data and show the results in Table \ref{tab:null}. 
 Our model shows consistent improvement and much lower variance compared to the simple classification baseline under different ratios of null instances. 

\begin{table}[t]
    \centering
    \small 
    \begin{tabular}{l  c c c }
    \toprule 
 Model   & 20\% & 50\%  & 100\% \\
      \midrule 
      \multicolumn{4}{c}{\textit{Identification}} \\
      CLS & 44.6 $\pm$ 4.5 &  45.3 $\pm$ 5.5 & 42.1 $\pm$ 4.8\\
    \ours   & 67.9 $\pm$ 1.3 & 64.7 $\pm$ 1.0  & 62.8 $\pm$ 1.0 \\
    \midrule 
    \multicolumn{4}{c}{\textit{Identification + Localization}} \\
    CLS  & 42.1 $\pm$ 4.1 & 43.5 $\pm$ 5.1 &  40.0 $\pm$ 4.4 \\
    \ours  & 65.4 $\pm$ 1.1 & 63.2 $\pm$ 0.9  & 60.4 $\pm$ 1.1 \\ 
     \bottomrule 
    \end{tabular}
    \caption{Few-shot event detection micro-F1 scores(\%) on MAVEN with 10-shot training examples under different NULL instance ratios. CLS refers to the setting where our identification model is replaced by a classification model using the \texttt{[CLS]} token as the sentence representation.}
    \label{tab:null}
\end{table}

\begin{table*}[t]
    \centering
    \small 
    \begin{tabular}{l l l| c c }
    \toprule 
      Id Method & Id Loss & Loc Method & $K=5$ & $K=10$  \\
     \midrule 
     Classification & CE & Attn-CRF & 18.88$\pm$ 2.6 &  44.03 $\pm$ 5.5 \\ 
     Prompt & Margin Loss & Attn-CRF & 58.62$\pm$1.4 & 68.59 $\pm$ 0.8 \\ 
      Prompt & ThresholdCE & Attn-CRF  & \textbf{60.64} $\pm$ 1.0 & \textbf{69.51} $\pm$ 1.5 \\
      \midrule 
      Gold & ThresholdCE & Attn-CRF & 79.55 $\pm$0.9 & 85.08 $\pm$ 0.9 \\
     \bottomrule 
    \end{tabular}
    \caption{Ablation studies on few-shot event detection for MAVEN. Reported results are micro-F1 scores(\%) with different number $K$ of training examples per type.}
    \label{tab:fewshot-additional}
\end{table*}

\begin{table}[ht]
    \centering
    \small 
    \begin{tabular}{l l |c c c}
    \toprule 
      Id Model & Loc Model  & Prec & Recall & F1  \\
       \midrule 
       $\checkmark$ & Attn-CRF  & 70.9 & 76.1 & 73.4 \\
       \midrule 
      $\checkmark$  & CRF & 68.3 & 74.9 & 71.5 \\
      $\checkmark$  & QA  & 72.5 & 69.0 & 70.7  \\
      $\checkmark$ & Span Classifier & 63.5 & 78.3 & 70.1 \\
      \midrule 
      Enumerate &  Attn-CRF & 54.5 & 81.3 & 65.3 \\
      Margin loss & Attn-CRF & 69.8 & 75.9 & 72.7 \\
      \bottomrule
    \end{tabular}
    \caption{Model ablations on ACE+.}
    \label{tab:ablations}
\end{table}
\subsection{Model Design Choices}
\label{sec:design-choices}
We design our identification model as a prompt-based model trained with NULL-thresholded cross entropy loss \textbf{ThresholdCE} (Equation \ref{eq:id-loss}) and our localization model as an attention-enhanced single-class CRF tagger \textbf{Attn-CRF}. %

For the identification model, we explore alternative modeling choices in Table \ref{tab:fewshot-additional}.
We compare with a classification model that takes the \texttt{[CLS]} token as the sentence representation and trained with cross entropy loss (\textbf{Classification +CE}). This method performs poorly as it is not able to take advantage of the language modeling capability as in the prompt-based formulation and the classification layer needs to be trained from scratch. We also consider changing our \textbf{ThresholdCE} loss to a margin-based ranking loss (\textbf{Prompt + Margin Loss}). This alternative loss function works relatively well, demonstrating that the prompt-based formulation is more critical than the loss function itself. The slight performance drop might be attributed to the fact the the margin loss treats each label equally, whereas our \textbf{ThresholdCE} loss puts more emphasis on positive labels.

In Table \ref{tab:ablations}, we further experiment with other designs in replacement of our localization model including the question answering (\textbf{QA}) formulation~\cite{du-cardie-2020-event,liu-etal-2020-event} , the span classification formulation (\textbf{Span Classifier}) and the vanilla \textbf{CRF} model. For the single-class CRF model, we remove the attention based early-interaction term in Equation 7.
In the question answering formulation, we compute the scores of the token being the first token in the answer (the answer head) and being the last token in the answer (the answer tail) separately. 
This simple QA model cannot handle multiple ``answers" per sentence, so we extend it to a span classification model where each span is scored independently and assigned a binary label.
Although the span classifier can handle multiple triggers in the same sentence, it suffers from low precision. Compared to the QA and span classifier variant, the vanilla CRF model benefits from modeling the correlation between neighboring tokens. Additionally, our attention-enhanced CRF layer can further improve upon the vanilla CRF model by 1.9 \% F1 points.
One alternative to our two-stage \textit{identify-then-localize} framework is to simply enumerate all possible event types and attempt to localize the trigger for each event (\textbf{Enumerate}).
As shown in the last row of Table \ref{tab:ablations}, this model has high recall at the cost of low precision. Additionally, with $N$ event types in the ontology, this model requires $N$ times training time and inference time.

\section{Related Work}
\paragraph{Prompt-Tuning}
In prompt-tuning the task specifications (task description or examples) are provided as part of the input. 
Depending on the format of the prompt, prompt-tuning methods can be divided into cloze-style prompts for classification~\cite{schick-schutze-2021-exploiting,schick-schutze-2021-just} and open-ended prompts for generation~\cite{li-liang-2021-prefix}. 

Application-wise, 
some recent attempts to apply prompt-tuning to informative extraction tasks include named entity recognition~\cite{Ning2021PromptEntityTyping} and relation extraction~\cite{Han2021PTRPT,Chen2022KnowPromptKP} but they focus on improving the classification component of these tasks. Our model, on the other hand, handles both multi-label classification (identification) task and the localization task.

\paragraph{Low Resource Event Detection}
Low resource event detection aims to alleviate the need for heavy annotation.
This includes settings such as zero-shot transfer learning~\cite{huang-etal-2018-zero, lyu-etal-2021-zero,liu-etal-2022-dynamic}, cross-lingual transfer~\cite{subburathinam-etal-2019-cross,huang-etal-2022-multilingual-generative}, keyword-based supervision~\cite{zhang-etal-2021-zero}, lifelong learning~\cite{Yu2021}, learning from annotation guideline~\cite{weaksupervision2022} and few-shot learning~\cite{peng-etal-2016-event,lai-etal-2020-extensively,shen-etal-2021-adaptive,cong-etal-2021-shot,Chen2021HoneyOP}.

In terms of methodology, prototype-based methods~\cite{Deng2020MetaLearningWD, zhang-etal-2021-zero, cong-etal-2021-shot,shen-etal-2021-adaptive} have been a popular choice. %
Another idea is to transfer knowledge from semantic parsers, such as AMR~\cite{wang-etal-2021-cleve,huang-etal-2018-zero} or SRL~\cite{zhang-etal-2021-zero, lyu-etal-2021-zero} parsers.
QA-based~\cite{du-cardie-2020-event,liu-etal-2020-event} and generation-based methods \cite{li-etal-2021-document,hsu2021degree,liu-etal-2022-dynamic} can also be adapted for few-shot event extraction and benefit from parameter-sharing since event type information can be incorporated into the question or generation prefix. %

\section{Conclusions and Future Work}
In this paper we propose \ours,~ an identify-and-localize framework for event detection. 
Specifically, we first \textit{identify} the event types present in the context and then find the trigger \textit{location} based on type-specific event knowledge. The identification stage benefits from the few-shot capability of prompt-tuning and the localization stage enable structured prediction with shared parameters over all types.
We show that this approach performs on par with SOTA in the fully supervised setting and outperforms existing methods for few-shot event detection, achieving 15\% absolute F1 score gain on FewEvent and 3\% gain on MAVEN.
We see that as the number of event types increases, the identification performance becomes a bottleneck, which calls for additional exploration for large event ontologies.

\newpage 
\section{Limitations}
Our model addresses the event detection task. We have limited our scope to sentence-level events and the datasets we use in this paper are derived from English news articles and Wikipedia.

To achieve the best results, our model requires a human-designed prompt and verbalizers (highly indicative words for each event type). While we, and our work such as \cite{Hu2021KnowledgeablePI,cui-etal-2022-prototypical} show the potential of using automatic verbalizers, it will still cause a slight degrade to the performance. To inject extra event knowledge during the localization stage, there should be some external resource that can be aligned to the event ontology, such as the annotation guidelines for ACE or FrameNet for MAVEN.

\section{Ethical Considerations}
Event detection is a standard component in the event extraction pipeline. We are not aware of any immediate research on bias in event extraction systems but bias has been reported for masked language models~\cite{nangia-etal-2020-crows} which we utilizing in our identification stage. 

Our paper does not present new datasets, but rather reuses the ACE, FewEvent and MAVEN datasets.

\bibliographystyle{acl_natbib}
\bibliography{anthology, custom}

\begin{thebibliography}{44}
\expandafter\ifx\csname natexlab\endcsname\relax\def\natexlab#1{#1}\fi

\bibitem[{Chen et~al.(2021)Chen, Lin, Han, and Sun}]{Chen2021HoneyOP}
Jiawei Chen, Hongyu Lin, Xianpei Han, and Le~Sun. 2021.
\newblock Honey or poison? solving the trigger curse in few-shot event
  detection via causal intervention.
\newblock \emph{EMNLP}.

\bibitem[{Chen et~al.(2022)Chen, Zhang, Zhang, Xie, Deng, Yao, Tan, Huang, Si,
  and Chen}]{Chen2022KnowPromptKP}
Xiang Chen, Ningyu Zhang, Ningyu Zhang, Xin Xie, Shumin Deng, Yunzhi Yao,
  Chuanqi Tan, Fei Huang, Luo Si, and Huajun Chen. 2022.
\newblock Knowprompt: Knowledge-aware prompt-tuning with synergistic
  optimization for relation extraction.
\newblock \emph{Proceedings of the ACM Web Conference 2022}.

\bibitem[{Cong et~al.(2021)Cong, Cui, Yu, Liu, Yubin, and
  Wang}]{cong-etal-2021-shot}
Xin Cong, Shiyao Cui, Bowen Yu, Tingwen Liu, Wang Yubin, and Bin Wang. 2021.
\newblock \href {https://doi.org/10.18653/v1/2021.findings-acl.3} {{F}ew-{S}hot
  {E}vent {D}etection with {P}rototypical {A}mortized {C}onditional {R}andom
  {F}ield}.
\newblock In \emph{Findings of the Association for Computational Linguistics:
  ACL-IJCNLP 2021}, pages 28--40, Online. Association for Computational
  Linguistics.

\bibitem[{Cui et~al.(2022)Cui, Hu, Ding, Huang, and
  Liu}]{cui-etal-2022-prototypical}
Ganqu Cui, Shengding Hu, Ning Ding, Longtao Huang, and Zhiyuan Liu. 2022.
\newblock \href {https://doi.org/10.18653/v1/2022.acl-long.483} {Prototypical
  verbalizer for prompt-based few-shot tuning}.
\newblock In \emph{Proceedings of the 60th Annual Meeting of the Association
  for Computational Linguistics (Volume 1: Long Papers)}, pages 7014--7024,
  Dublin, Ireland. Association for Computational Linguistics.

\bibitem[{Deng et~al.(2020)Deng, Zhang, Kang, Zhang, Zhang, and
  Chen}]{Deng2020MetaLearningWD}
Shumin Deng, Ningyu Zhang, Jiaojian Kang, Yichi Zhang, Wei Zhang, and Huajun
  Chen. 2020.
\newblock Meta-learning with dynamic-memory-based prototypical network for
  few-shot event detection.
\newblock \emph{Proceedings of the 13th International Conference on Web Search
  and Data Mining}.

\bibitem[{Devlin et~al.(2019)Devlin, Chang, Lee, and
  Toutanova}]{devlin-etal-2019-bert}
Jacob Devlin, Ming-Wei Chang, Kenton Lee, and Kristina Toutanova. 2019.
\newblock \href {https://doi.org/10.18653/v1/N19-1423} {{BERT}: Pre-training of
  deep bidirectional transformers for language understanding}.
\newblock In \emph{Proceedings of the 2019 Conference of the North {A}merican
  Chapter of the Association for Computational Linguistics: Human Language
  Technologies, Volume 1 (Long and Short Papers)}, pages 4171--4186,
  Minneapolis, Minnesota. Association for Computational Linguistics.

\bibitem[{Ding et~al.(2021)Ding, Chen, Han, Xu, Xie, Zheng, Liu, Li, and
  Kim}]{Ning2021PromptEntityTyping}
Ning Ding, Yulin Chen, Xu~Han, Guangwei Xu, Pengjun Xie, Hai-Tao Zheng, Zhiyuan
  Liu, Juan-Zi Li, and Hong-Gee Kim. 2021.
\newblock Prompt-learning for fine-grained entity typing.
\newblock \emph{ArXiv}, abs/2108.10604.

\bibitem[{Du and Cardie(2020)}]{du-cardie-2020-event}
Xinya Du and Claire Cardie. 2020.
\newblock \href {https://doi.org/10.18653/v1/2020.emnlp-main.49} {Event
  extraction by answering (almost) natural questions}.
\newblock In \emph{Proceedings of the 2020 Conference on Empirical Methods in
  Natural Language Processing (EMNLP)}, pages 671--683, Online. Association for
  Computational Linguistics.

\bibitem[{Gao et~al.(2019)Gao, Han, Zhu, Liu, Li, Sun, and
  Zhou}]{gao-etal-2019-fewrel}
Tianyu Gao, Xu~Han, Hao Zhu, Zhiyuan Liu, Peng Li, Maosong Sun, and Jie Zhou.
  2019.
\newblock \href {https://doi.org/10.18653/v1/D19-1649} {{F}ew{R}el 2.0: Towards
  more challenging few-shot relation classification}.
\newblock In \emph{Proceedings of the 2019 Conference on Empirical Methods in
  Natural Language Processing and the 9th International Joint Conference on
  Natural Language Processing (EMNLP-IJCNLP)}, pages 6250--6255, Hong Kong,
  China. Association for Computational Linguistics.

\bibitem[{Han et~al.(2021)Han, Zhao, Ding, Liu, and Sun}]{Han2021PTRPT}
Xu~Han, Weilin Zhao, Ning Ding, Zhiyuan Liu, and Maosong Sun. 2021.
\newblock Ptr: Prompt tuning with rules for text classification.
\newblock \emph{ArXiv}, abs/2105.11259.

\bibitem[{Hsu et~al.(2021)Hsu, Huang, Boschee, Miller, Natarajan, Chang, and
  Peng}]{hsu2021degree}
I-Hung Hsu, Kuan-Hao Huang, Elizabeth Boschee, Scott Miller, Prem Natarajan,
  Kai-Wei Chang, and Nanyun Peng. 2021.
\newblock \href {http://arxiv.org/abs/2108.12724} {Degree: A data-efficient
  generative event extraction model}.

\bibitem[{Hu et~al.(2021)Hu, Ding, Wang, Liu, Li, and
  Sun}]{Hu2021KnowledgeablePI}
Shengding Hu, Ning Ding, Huadong Wang, Zhiyuan Liu, Juan-Zi Li, and Maosong
  Sun. 2021.
\newblock Knowledgeable prompt-tuning: Incorporating knowledge into prompt
  verbalizer for text classification.
\newblock \emph{ArXiv}, abs/2108.02035.

\bibitem[{Huang et~al.(2022)Huang, Hsu, Natarajan, Chang, and
  Peng}]{huang-etal-2022-multilingual-generative}
Kuan-Hao Huang, I-Hung Hsu, Prem Natarajan, Kai-Wei Chang, and Nanyun Peng.
  2022.
\newblock \href {https://doi.org/10.18653/v1/2022.acl-long.317} {Multilingual
  generative language models for zero-shot cross-lingual event argument
  extraction}.
\newblock In \emph{Proceedings of the 60th Annual Meeting of the Association
  for Computational Linguistics (Volume 1: Long Papers)}, pages 4633--4646,
  Dublin, Ireland. Association for Computational Linguistics.

\bibitem[{Huang et~al.(2018)Huang, Ji, Cho, Dagan, Riedel, and
  Voss}]{huang-etal-2018-zero}
Lifu Huang, Heng Ji, Kyunghyun Cho, Ido Dagan, Sebastian Riedel, and Clare
  Voss. 2018.
\newblock \href {https://doi.org/10.18653/v1/P18-1201} {Zero-shot transfer
  learning for event extraction}.
\newblock In \emph{Proceedings of the 56th Annual Meeting of the Association
  for Computational Linguistics (Volume 1: Long Papers)}, pages 2160--2170,
  Melbourne, Australia. Association for Computational Linguistics.

\bibitem[{Ji and Grishman(2008)}]{ji2008refining}
Heng Ji and Ralph Grishman. 2008.
\newblock Refining event extraction through unsupervised cross-document
  inference.
\newblock In \emph{In Proceedings of the Annual Meeting of the Association of
  Computational Linguistics (ACL 2008). Ohio, USA}.

\bibitem[{Ji and Grishman(2011)}]{ji2011knowledge}
Heng Ji and Ralph Grishman. 2011.
\newblock Knowledge base population: Successful approaches and challenges.
\newblock In \emph{Proc. ACL2011}.

\bibitem[{Lai et~al.(2020)Lai, Nguyen, and
  Dernoncourt}]{lai-etal-2020-extensively}
Viet~Dac Lai, Thien~Huu Nguyen, and Franck Dernoncourt. 2020.
\newblock \href {https://doi.org/10.18653/v1/2020.nuse-1.5} {Extensively
  matching for few-shot learning event detection}.
\newblock In \emph{Proceedings of the First Joint Workshop on Narrative
  Understanding, Storylines, and Events}, pages 38--45, Online. Association for
  Computational Linguistics.

\bibitem[{Li et~al.(2021)Li, Ji, and Han}]{li-etal-2021-document}
Sha Li, Heng Ji, and Jiawei Han. 2021.
\newblock \href {https://doi.org/10.18653/v1/2021.naacl-main.69}
  {Document-level event argument extraction by conditional generation}.
\newblock In \emph{Proceedings of the 2021 Conference of the North American
  Chapter of the Association for Computational Linguistics: Human Language
  Technologies}, pages 894--908, Online. Association for Computational
  Linguistics.

\bibitem[{Li and Liang(2021)}]{li-liang-2021-prefix}
Xiang~Lisa Li and Percy Liang. 2021.
\newblock \href {https://doi.org/10.18653/v1/2021.acl-long.353} {Prefix-tuning:
  Optimizing continuous prompts for generation}.
\newblock In \emph{Proceedings of the 59th Annual Meeting of the Association
  for Computational Linguistics and the 11th International Joint Conference on
  Natural Language Processing (Volume 1: Long Papers)}, pages 4582--4597,
  Online. Association for Computational Linguistics.

\bibitem[{Lin et~al.(2020)Lin, Ji, Huang, and Wu}]{lin-etal-2020-joint}
Ying Lin, Heng Ji, Fei Huang, and Lingfei Wu. 2020.
\newblock \href {https://doi.org/10.18653/v1/2020.acl-main.713} {A joint neural
  model for information extraction with global features}.
\newblock In \emph{Proceedings of the 58th Annual Meeting of the Association
  for Computational Linguistics}, pages 7999--8009, Online. Association for
  Computational Linguistics.

\bibitem[{Liu et~al.(2020)Liu, Chen, Liu, Bi, and Liu}]{liu-etal-2020-event}
Jian Liu, Yubo Chen, Kang Liu, Wei Bi, and Xiaojiang Liu. 2020.
\newblock \href {https://doi.org/10.18653/v1/2020.emnlp-main.128} {Event
  extraction as machine reading comprehension}.
\newblock In \emph{Proceedings of the 2020 Conference on Empirical Methods in
  Natural Language Processing (EMNLP)}, pages 1641--1651, Online. Association
  for Computational Linguistics.

\bibitem[{Liu et~al.(2022)Liu, Huang, Shi, and Wang}]{liu-etal-2022-dynamic}
Xiao Liu, Heyan Huang, Ge~Shi, and Bo~Wang. 2022.
\newblock \href {https://doi.org/10.18653/v1/2022.acl-long.358} {Dynamic
  prefix-tuning for generative template-based event extraction}.
\newblock In \emph{Proceedings of the 60th Annual Meeting of the Association
  for Computational Linguistics (Volume 1: Long Papers)}, pages 5216--5228,
  Dublin, Ireland. Association for Computational Linguistics.

\bibitem[{Liu et~al.(2019)Liu, Ott, Goyal, Du, Joshi, Chen, Levy, Lewis,
  Zettlemoyer, and Stoyanov}]{Liu2019RoBERTaAR}
Yinhan Liu, Myle Ott, Naman Goyal, Jingfei Du, Mandar Joshi, Danqi Chen, Omer
  Levy, Mike Lewis, Luke Zettlemoyer, and Veselin Stoyanov. 2019.
\newblock Roberta: A robustly optimized bert pretraining approach.
\newblock \emph{ArXiv}, abs/1907.11692.

\bibitem[{Lu et~al.(2021)Lu, Lin, Xu, Han, Tang, Li, Sun, Liao, and
  Chen}]{lu-etal-2021-text2event}
Yaojie Lu, Hongyu Lin, Jin Xu, Xianpei Han, Jialong Tang, Annan Li, Le~Sun,
  Meng Liao, and Shaoyi Chen. 2021.
\newblock \href {https://doi.org/10.18653/v1/2021.acl-long.217}
  {{T}ext2{E}vent: Controllable sequence-to-structure generation for end-to-end
  event extraction}.
\newblock In \emph{Proceedings of the 59th Annual Meeting of the Association
  for Computational Linguistics and the 11th International Joint Conference on
  Natural Language Processing (Volume 1: Long Papers)}, pages 2795--2806,
  Online. Association for Computational Linguistics.

\bibitem[{Lyu et~al.(2021)Lyu, Zhang, Sulem, and Roth}]{lyu-etal-2021-zero}
Qing Lyu, Hongming Zhang, Elior Sulem, and Dan Roth. 2021.
\newblock \href {https://doi.org/10.18653/v1/2021.acl-short.42} {Zero-shot
  event extraction via transfer learning: {C}hallenges and insights}.
\newblock In \emph{Proceedings of the 59th Annual Meeting of the Association
  for Computational Linguistics and the 11th International Joint Conference on
  Natural Language Processing (Volume 2: Short Papers)}, pages 322--332,
  Online. Association for Computational Linguistics.

\bibitem[{Nangia et~al.(2020)Nangia, Vania, Bhalerao, and
  Bowman}]{nangia-etal-2020-crows}
Nikita Nangia, Clara Vania, Rasika Bhalerao, and Samuel~R. Bowman. 2020.
\newblock \href {https://doi.org/10.18653/v1/2020.emnlp-main.154}
  {{C}row{S}-pairs: A challenge dataset for measuring social biases in masked
  language models}.
\newblock In \emph{Proceedings of the 2020 Conference on Empirical Methods in
  Natural Language Processing (EMNLP)}, pages 1953--1967, Online. Association
  for Computational Linguistics.

\bibitem[{Nguyen et~al.(2021)Nguyen, Lai, and Nguyen}]{nguyen-etal-2021-cross}
Minh~Van Nguyen, Viet Lai, and Thien~Huu Nguyen. 2021.
\newblock \href {https://doi.org/10.18653/v1/2021.naacl-main.3} {Cross-task
  instance representation interactions and label dependencies for joint
  information extraction with graph convolutional networks}.
\newblock In \emph{Proceedings of the 2021 Conference of the North American
  Chapter of the Association for Computational Linguistics: Human Language
  Technologies}, pages 27--38, Online. Association for Computational
  Linguistics.

\bibitem[{Peng et~al.(2016)Peng, Song, and Roth}]{peng-etal-2016-event}
Haoruo Peng, Yangqiu Song, and Dan Roth. 2016.
\newblock \href {https://doi.org/10.18653/v1/D16-1038} {Event detection and
  co-reference with minimal supervision}.
\newblock In \emph{Proceedings of the 2016 Conference on Empirical Methods in
  Natural Language Processing}, pages 392--402, Austin, Texas. Association for
  Computational Linguistics.

\bibitem[{Sabo et~al.(2021)Sabo, Elazar, Goldberg, and
  Dagan}]{Sabo2021RevisitingFR}
O.~Mahamane~Sani Sabo, Yanai Elazar, Yoav Goldberg, and Ido Dagan. 2021.
\newblock Revisiting few-shot relation classification: Evaluation data and
  classification schemes.
\newblock \emph{Transactions of the Association for Computational Linguistics},
  9:691--706.

\bibitem[{Schick and
  Sch{\"u}tze(2021{\natexlab{a}})}]{schick-schutze-2021-exploiting}
Timo Schick and Hinrich Sch{\"u}tze. 2021{\natexlab{a}}.
\newblock \href {https://aclanthology.org/2021.eacl-main.20} {Exploiting
  cloze-questions for few-shot text classification and natural language
  inference}.
\newblock In \emph{Proceedings of the 16th Conference of the European Chapter
  of the Association for Computational Linguistics: Main Volume}, pages
  255--269, Online. Association for Computational Linguistics.

\bibitem[{Schick and
  Sch{\"u}tze(2021{\natexlab{b}})}]{schick-schutze-2021-just}
Timo Schick and Hinrich Sch{\"u}tze. 2021{\natexlab{b}}.
\newblock \href {https://doi.org/10.18653/v1/2021.naacl-main.185} {It{'}s not
  just size that matters: Small language models are also few-shot learners}.
\newblock In \emph{Proceedings of the 2021 Conference of the North American
  Chapter of the Association for Computational Linguistics: Human Language
  Technologies}, pages 2339--2352, Online. Association for Computational
  Linguistics.

\bibitem[{Shen et~al.(2021)Shen, Wu, Qi, Li, Haffari, and
  Bi}]{shen-etal-2021-adaptive}
Shirong Shen, Tongtong Wu, Guilin Qi, Yuan-Fang Li, Gholamreza Haffari, and
  Sheng Bi. 2021.
\newblock \href {https://doi.org/10.18653/v1/2021.findings-acl.214} {Adaptive
  knowledge-enhanced {B}ayesian meta-learning for few-shot event detection}.
\newblock In \emph{Findings of the Association for Computational Linguistics:
  ACL-IJCNLP 2021}, pages 2417--2429, Online. Association for Computational
  Linguistics.

\bibitem[{Snell et~al.(2017)Snell, Swersky, and
  Zemel}]{Snell2017PrototypicalNF}
Jake Snell, Kevin Swersky, and Richard~S. Zemel. 2017.
\newblock Prototypical networks for few-shot learning.
\newblock In \emph{NIPS}.

\bibitem[{Subburathinam et~al.(2019)Subburathinam, Lu, Ji, May, Chang, Sil, and
  Voss}]{subburathinam-etal-2019-cross}
Ananya Subburathinam, Di~Lu, Heng Ji, Jonathan May, Shih-Fu Chang, Avirup Sil,
  and Clare Voss. 2019.
\newblock \href {https://doi.org/10.18653/v1/D19-1030} {Cross-lingual structure
  transfer for relation and event extraction}.
\newblock In \emph{Proceedings of the 2019 Conference on Empirical Methods in
  Natural Language Processing and the 9th International Joint Conference on
  Natural Language Processing (EMNLP-IJCNLP)}, pages 313--325, Hong Kong,
  China. Association for Computational Linguistics.

\bibitem[{Sung et~al.(2018)Sung, Yang, Zhang, Xiang, Torr, and
  Hospedales}]{Sung2018RelNet}
Flood Sung, Yongxin Yang, Li~Zhang, Tao Xiang, Philip H.~S. Torr, and
  Timothy~M. Hospedales. 2018.
\newblock Learning to compare: Relation network for few-shot learning.
\newblock \emph{2018 IEEE/CVF Conference on Computer Vision and Pattern
  Recognition}, pages 1199--1208.

\bibitem[{Vinyals et~al.(2016)Vinyals, Blundell, Lillicrap, Kavukcuoglu, and
  Wierstra}]{Vinyals2016MatchingNF}
Oriol Vinyals, Charles Blundell, Timothy~P. Lillicrap, Koray Kavukcuoglu, and
  Daan Wierstra. 2016.
\newblock Matching networks for one shot learning.
\newblock In \emph{NIPS}.

\bibitem[{Wadden et~al.(2019)Wadden, Wennberg, Luan, and
  Hajishirzi}]{wadden-etal-2019-entity}
David Wadden, Ulme Wennberg, Yi~Luan, and Hannaneh Hajishirzi. 2019.
\newblock \href {https://doi.org/10.18653/v1/D19-1585} {Entity, relation, and
  event extraction with contextualized span representations}.
\newblock In \emph{Proceedings of the 2019 Conference on Empirical Methods in
  Natural Language Processing and the 9th International Joint Conference on
  Natural Language Processing (EMNLP-IJCNLP)}, pages 5784--5789, Hong Kong,
  China. Association for Computational Linguistics.

\bibitem[{Wang et~al.(2020)Wang, Wang, Han, Jiang, Han, Liu, Li, Li, Lin, and
  Zhou}]{wang-etal-2020-maven}
Xiaozhi Wang, Ziqi Wang, Xu~Han, Wangyi Jiang, Rong Han, Zhiyuan Liu, Juanzi
  Li, Peng Li, Yankai Lin, and Jie Zhou. 2020.
\newblock \href {https://doi.org/10.18653/v1/2020.emnlp-main.129} {{MAVEN}: {A}
  {M}assive {G}eneral {D}omain {E}vent {D}etection {D}ataset}.
\newblock In \emph{Proceedings of the 2020 Conference on Empirical Methods in
  Natural Language Processing (EMNLP)}, pages 1652--1671, Online. Association
  for Computational Linguistics.

\bibitem[{Wang et~al.(2021)Wang, Wang, Han, Lin, Hou, Liu, Li, Li, and
  Zhou}]{wang-etal-2021-cleve}
Ziqi Wang, Xiaozhi Wang, Xu~Han, Yankai Lin, Lei Hou, Zhiyuan Liu, Peng Li,
  Juanzi Li, and Jie Zhou. 2021.
\newblock \href {https://doi.org/10.18653/v1/2021.acl-long.491} {{CLEVE}:
  {C}ontrastive {P}re-training for {E}vent {E}xtraction}.
\newblock In \emph{Proceedings of the 59th Annual Meeting of the Association
  for Computational Linguistics and the 11th International Joint Conference on
  Natural Language Processing (Volume 1: Long Papers)}, pages 6283--6297,
  Online. Association for Computational Linguistics.

\bibitem[{Yang and Katiyar(2020)}]{yang-katiyar-2020-simple}
Yi~Yang and Arzoo Katiyar. 2020.
\newblock \href {https://doi.org/10.18653/v1/2020.emnlp-main.516} {Simple and
  effective few-shot named entity recognition with structured nearest neighbor
  learning}.
\newblock In \emph{Proceedings of the 2020 Conference on Empirical Methods in
  Natural Language Processing (EMNLP)}, pages 6365--6375, Online. Association
  for Computational Linguistics.

\bibitem[{Yu et~al.(2021)Yu, Ji, and Natarajan}]{Yu2021}
Pengfei Yu, Heng Ji, and Premkumar Natarajan. 2021.
\newblock Lifelong event detection with knowledge transfer.
\newblock In \emph{Proc. The 2021 Conference on Empirical Methods in Natural
  Language Processing (EMNLP2021)}.

\bibitem[{Yu et~al.(2022)Yu, Zhang, Voss, May, and Ji}]{weaksupervision2022}
Pengfei Yu, Zixuan Zhang, Clare Voss, Jonathan May, and Heng Ji. 2022.
\newblock Event extractor with only a few examples.
\newblock In \emph{Proc. NAACL2022 workshop on Deep Learning for Low Resource
  NLP}.

\bibitem[{Zhang et~al.(2021)Zhang, Wang, and Roth}]{zhang-etal-2021-zero}
Hongming Zhang, Haoyu Wang, and Dan Roth. 2021.
\newblock \href {https://doi.org/10.18653/v1/2021.findings-acl.114}
  {{Z}ero-shot {L}abel-aware {E}vent {T}rigger and {A}rgument
  {C}lassification}.
\newblock In \emph{Findings of the Association for Computational Linguistics:
  ACL-IJCNLP 2021}, pages 1331--1340, Online. Association for Computational
  Linguistics.

\bibitem[{Zhou et~al.(2021)Zhou, Huang, Ma, and
  Huang}]{Zhou2021AdaptiveThreshold}
Wenxuan Zhou, Kevin Huang, Tengyu Ma, and Jinke Huang. 2021.
\newblock Document-level relation extraction with adaptive thresholding and
  localized context pooling.
\newblock In \emph{AAAI}.

\end{thebibliography}

\appendix 
\section{Dataset Details}
\label{sec:dataset}
FewEvent is designed to be a few-shot event detection benchmark aggregating data from ACE, TAC-KBP~\cite{ji2011knowledge} and expanding to additional event types related to sports, music, education, etc. from Wikipedia and Freebase. 
We follow the data split released by \cite{cong-etal-2021-shot}.
In the data provided, sentences are organized by event type and each sentence only has one event mention annotation.

MAVEN is the largest human-annotated event detection dataset to date, covering 4,480 documents and 168 event types. We use MAVEN for the few-shot setting following \cite{Chen2021HoneyOP}. 

ACE2005 is the most widely used dataset for event extraction. 
For data preprocessing, we follow \cite{lin-etal-2020-joint} and keep multi-word triggers and pronouns. We denote this version of ACE2005 as ACE+.  Since FewEvent has significant data overlap with ACE2005, we do not further experiment with the few-shot setting on ACE 2005.

\begin{table}[ht]
    \centering
    \small 
    \begin{tabular}{l|c c c }
    \toprule 
         & Train & Dev & Test \\
         \midrule 
        \# Types & 80 & 10 & 10 \\
        \# Sents & 67,982 & 2,173 & 697 \\
    \bottomrule 
    \end{tabular}
    \caption{Data split for FewEvent.}
    \label{tab:fewevent-split}
\end{table}
In the N-way K-shot experiments, we randomly sample $N$ event types from the test set and then sample $K$ labeled instances of that event type for training.

For MAVEN, we follow the data split by \cite{Chen2021HoneyOP} and use the sentences containing the most frequent 120 event types as the training set.
The sentences containing the remaining 45 event type are then split into half as the dev and test set. We use the same random seed as \cite{Chen2021HoneyOP} to ensure the same split.
\begin{table}[ht]
    \centering
    \small 
    \begin{tabular}{l|c c c }
    \toprule 
         & Train & Dev & Test \\
         \midrule 
        \# Types & 125 & 45 & 45 \\
        \# Sents &  86, 551 & 1,532 & 1,555 \\
        \# Events & 287, 516 & 1,741 & 1,806 \\
    \bottomrule 
    \end{tabular}
    \caption{Data split for MAVEN.}
    \label{tab:maven-split}
\end{table}

For ACE, we use the data split in \cite{lin-etal-2020-joint}. The same 33 event types are shared in the training, dev and test set.
\begin{table}[ht]
    \centering
    \small 
    \begin{tabular}{l|c c c }
    \toprule 
         & Train & Dev & Test \\
         \midrule 
        \# Sents &  19, 240 & 902 & 676 \\
        \# Events & 4,419 & 468 & 424 \\
    \bottomrule 
    \end{tabular}
    \caption{Data split for ACE+.}
    \label{tab:ace-split}
\end{table}

\section{Model Hyperparameters} 

For the experiments on ACE+, we used the settings and hyperparameters as shown in Table \ref{tab:ace_param}.

\begin{table}[ht]
    \centering
    \small 
    \begin{tabular}{l|c}
    \toprule 
         Parameter &  Value  \\
         \midrule 
        Encoder & Roberta-large \\
        Max seq len & 200 \\
        Batch size & 16 \\
        Learning rate & $1e-5$ \\
        Learning rate schedule & Linear \\
        Weight decay & $1e-5$ \\
        Warmup steps & 1000 \\
        Epochs & 10 \\
        Adam $\epsilon$ & $1e-8$ \\
        Gradient clipping & 1.0 \\
        \bottomrule 
    \end{tabular}
    \caption{ACE+ hyperparameters}
    \label{tab:ace_param}
\end{table}

For all few-shot experiments, we use the parameters listed in Table \ref{tab:fewshot_param}.
Experiments were run on a single Nvidia RTX A6000 GPU or 3080 GPU.
Our BERT-based model contains 134.2M parameters, among which 133M parameters are from the bert-base-uncased model with a LM head.
Since our model is only trained on $K$ examples for the test types, a single run (training and testing) takes only about 10 minutes.

\begin{table}[ht]
    \centering
    \small 
    \begin{tabular}{l|c}
    \toprule 
         Parameter &  Value  \\
         \midrule 
        Encoder & bert-base-uncased \\
        Max seq len & 200 \\
        Batch size & 8 \\
        Learning rate & $2e-5$ \\
        Learning rate schedule & Linear \\
        Weight decay & $1e-5$ \\
        Warmup steps & 0 \\
        Epochs & 20 \\
        Adam $\epsilon$ & $1e-8$ \\
        Gradient clipping & 1.0 \\
        \bottomrule 
    \end{tabular}
    \caption{Few-shot experiment hyperparameters.}
    \label{tab:fewshot_param}
\end{table}
\subsection{Injecting Event Knowledge}
In our model, event knowledge is present in the \textit{verbalizer} in the identification stage and the \textit{type-aware prompt} in the localization stage.
For the type-aware prompt, in our pilot experiments, we considered using the event definition or event keywords and compare it against the baseline of using the filled prompt from the identification stage. As seen in Table \ref{tab:loc-prompt}, the event verbalizer alone is relatively informative and adding more event keywords from the lexical units can provide a minor gain.
Definitions from FrameNet are highly abstract, which may undermine their value in assisting event detection. In the next Appendix section \ref{sec:multi_verb}, we further explore using more verbalizers to enhance event detection performance.

\begin{table}[t]
    \centering
    \small 
    \begin{tabular}{l|c c  }
    \toprule 
     Event knowledge   &  Id F1 & Loc F1 \\
     \midrule 
       Verbalizer   & 64.8 $\pm$ 1.3 & 62.0 $\pm$ 1.5 \\
       Verbalizer + Definition & 64.8 $\pm$ 1.3 & 62.3 $\pm$ 1.5 \\
       Verbalizer + Keywords & \textbf{65.5} $\pm$ 1.1 & \textbf{63.1} $\pm$ 1.1 \\
       \bottomrule
    \end{tabular}
    \caption{Comparison of using different types of event knowledge to construct the type-aware prompt for localization~(RoBERTa-base model). The event verbalizer is present in the filled prompt. We use at most 3 keywords per event type.
    }
    \label{tab:loc-prompt}
\end{table}

\section{Using Multiple Verbalizers}
\label{sec:multi_verb}
In the previous experiments, we use one manually selected verbalizer per event type. A natural question is whether more verbalizers will help. We use MAVEN for this set of experiments since MAVEN provides alignments between its event types and FrameNet frames. The FrameNet\footnote{\url{https://framenet.icsi.berkeley.edu/fndrupal/frameIndex}} definitions and lexical units can then serve as event knowledge.

When more than one verbalizer is used, we need to aggregate the scores over the verbalizer set. We experiment with 4 different types of aggregation operators: \texttt{avg}, \texttt{max}, \texttt{logsumexp}, and \texttt{weighted-avg}.
The \texttt{logsumexp} operator can be seen as a smoothed version of the \texttt{max} operator.
In the \texttt{weighted-avg} operator, the weights of the verbalizers are additional learnable parameters~\cite{Hu2021KnowledgeablePI}. As shown in Table \ref{tab:multi-verbalizer}, in the few-shot setting, using multiple verbalizers can provide 1.5-2\% F1 improvement on identification which translates to 1.6-2.2\% F1 improvement on the event detection task. In terms of aggregation methods, the \texttt{avg} operator is a simple and reliable choice with the best performance and lowest variance. Although the \texttt{wavg} operator is more expressive, it is hard to learn good weights with only 5 examples per event type.

\begin{table}[t]
    \centering
    \small 
    \begin{tabular}{l | c c }
    \toprule 
     Agg method &  Id F1 & Id+Loc F1 \\
      \midrule 
      \texttt{avg} & \textbf{67.5} $\pm$ 1.6 & \textbf{65.3} $\pm$ 1.4 \\ 
      \texttt{max} & 67.0 $\pm$ 2.2 & 64.7 $\pm$ 2.2\\
      \texttt{logsumexp} & 67.0 $\pm$ 1.9 & 64.7 $\pm$ 1.9 \\
      \texttt{wavg} & 67.4 $\pm$ 1.6 & 64.9 $\pm$ 1.7 \\
      \bottomrule
    \end{tabular}
    \caption{Using multiple verbalizers for the 45-way-5-shot event detection on the MAVEN dataset (RoBERTa-base model). To balance between frames that have different number of lexical units, we use at most 3 verbalizers. \texttt{wavg} stands for \texttt{weighted-avg}.
    }
    \label{tab:multi-verbalizer}
\end{table}

\section{Discussion on Few-shot Event Datasets}
\label{sec:appendix-dataset}
Few-shot learning for event detection was largely inspired by the few-shot classification work in computer vision literature~\cite{Vinyals2016MatchingNF,Snell2017PrototypicalNF, Sung2018RelNet} which assumes that images are sampled independently under the N-way K-shot setting. 
However, this assumption does not directly apply to context-dependent tasks such as event detection: the distribution of event types heavily depends on the document and is far from i.i.d. in practice.
This sampling procedure also leads to the absence of the NULL class (sentences without any event mentions), which is often abundant in real documents.

This data discrepancy has received some attention in other tasks such as relation extraction~\cite{gao-etal-2019-fewrel, Sabo2021RevisitingFR} but is under-explored for event detection. 
For example, FewEvent instances only contain one event type per sentence and do not include NULL class examples. Sentences from MAVEN may contain multiple event types but also exclude the case of NULL.
Thus, many previous works in few-shot event detection simply design their model to be a $K$-way classifier. ACE, the dataset which we use for supervised event detection, contains all these cases and the events follow a natural distribution but the small number of event types makes it less attractive to use as a few-shot benchmark.
Our model \ours~ is capable of handling these cases, as exemplified by our performance on ACE, but such abilities were not put to test on the current few-shot datasets.
As a result, we would like to remind readers of the possible inflation of few-shot performance on current benchmarks and call for future research on setting up better evaluation.

\end{document}